\documentclass{article}
\usepackage[utf8]{inputenc}
\usepackage{amsmath}
\usepackage{hyperref}
\usepackage{graphicx}
\usepackage{float}
\usepackage[T1]{fontenc}
\usepackage[french]{babel}
\usepackage{csquotes}
\usepackage[utf8]{inputenc}
\usepackage[table]{xcolor}
\usepackage{caption}
\usepackage{array}        
\usepackage{multirow}     
\usepackage{colortbl}     
\usepackage{array}
\usepackage{etoolbox}
\AtBeginEnvironment{tabular}{\centering}

\addto\captionsfrench{}
\addto\captionsfrench{}
\title{The Process of Categorical Clipping at the Core of the Genesis of Concepts in Synthetic Neural Cognition}

\author{
    Michael Pichat\thanks{Neocognition (Chrysippe R\&D) \& University of Paris \& Free Faculties of Philosophy and Psychology of Paris (ER IPC)} \\
    William Pogrund\thanks{Neocognition (Chrysippe R\&D) \& National Polytechnic Institute - Phelma, Grenoble Alpes University} \\
    Armanush Gasparian\thanks{Neocognition (Chrysippe R\&D)} \\
    Paloma Pichat\thanks{Neocognition (Chrysippe R\&D) \& Faculty of Medicine of Lyon East, University Lyon 1} \\
    Samuel Demarchi\thanks{Neocognition (Chrysippe R\&D) \& Department of Psychology, University of Paris 8} \\
    Michael Veillet-Guillem\thanks{Neocognition (Chrysippe R\&D) and Epitech Paris} \\
    Martin Corbet\textsuperscript{*} \\
    Théo Dasilva\textsuperscript{*}
}

\date{}

\begin{document}

\maketitle

\begin{abstract}
This article investigates, within the field of neuropsychology of artificial intelligence, the process of categorical segmentation performed by language models. This process involves, across different neural layers, the creation of new functional categorical dimensions to analyze the input textual data and perform the required tasks. Each neuron in a multilayer perceptron (MLP) network is associated with a specific category, generated by three factors carried by the neural aggregation function: categorical priming, categorical attention, and categorical phasing. At each new layer, these factors govern the formation of new categories derived from the categories of precursor neurons. Through a process of categorical clipping, these new categories are created by selectively extracting specific subdimensions from the preceding categories, constructing a distinction between a form and a categorical background. We explore several cognitive characteristics of this synthetic clipping in an exploratory manner: categorical reduction, categorical selectivity, separation of initial embedding dimensions, and segmentation of categorical zones.

\end{abstract}

\section{Introduction}
Synthetic categorical segmentation is the activity of formal neurons consisting of "cutting," within the token world to which a language model is exposed, a dimension of features used to analyze and categorize these tokens—a concept-in-act deemed relevant given the nature of the task to be performed by this model, as Vergnaud would argue \cite{Vergnaud2009, Vergnaud2016} in the field of cognitive psychology of human development. The result of this categorical segmentation is reflected in the creation, by each neuron, of a synthetic category of thought, a concept, or, to put it differently, a categorical dimension carried by this neuron \cite{Pichat2023, Pichat2024a}. This synthetic conceptual category is, among other things, defined by its extension, that is, the set of tokens for which the neuron associated with this category is (sufficiently) activated.

In a previous work \cite{Pichat2024d}, we investigated the mathematical-cognitive factors of categorical segmentation performed by the synthetic neurons of language models. In this preliminary exploratory study, we examined, both quantitatively and qualitatively, the genetic elements influencing this categorical segmentation. Based on the aggregation function \footnote{Bills et al. (2023), whose data we will use in this present study from their associated GitHub repository, provide a list "of the upstream and downstream neurons with the most positive and negative connections"; they define this list operationally as follows: "Definition of connection weights: neuron-neuron: for two neurons (l1, n1) and (l2, n2) with l1 < l2, the connection strength is defined as $h\{l1\}.mlp.c\_proj.w[:, n1, :] \ @ \ \text{diag}(h\{l2\}.ln\_2.g) \ @ \ h\{l2\}.mlp.c\_fc.w[:, :, n2]$." This list defines, within the dense layers (i.e., fully connected layers) of GPT-2XL, the weights through which each neuron in a target layer $n+1$ is connected to all the neurons of the previous layer $n$. It is based on these weights that the linear aggregation functions, referenced in this article, operate.}
 $\Sigma(w_{i,j} x_{i,j}) + b$, which partly governs this cognitive process, we identified three key causal elements of a mathematical and cognitive nature involved in this conceptual partitioning process.

First, the "x effect" or synthetic categorical priming, which refers to the fact that the activation of the categories carried by precursor neurons in layer $n$ affects the activation of the categories specific to their associated target neurons in layer $n+1$, thereby directly impacting their categorical extension. In other words, the more a token belongs to the extension of a precursor category in layer $n$ (i.e., the more this token is activated in the involved neuron), the greater its potential to belong to the extension of its superordinate category (i.e., its potential activation in the neuron of layer $n+1$). This phenomenon of categorical priming thus partly governs the categorical segmentation performed in layer $n+1$, that is, the determination of the subset of tokens constituting the categorical extension of the concepts carried by the neurons in layer $n+1$.

Second, the "w effect" or synthetic categorical attention, which relates to the fact that the weights of the connections between a target neuron (layer $n+1$) and its precursor neurons (layer $n$) govern the degree of relevance attributed to the precursor categories in constructing the categorical segment of their corresponding target neurons. This manifests qualitatively as a process of categorical complementation, genetically consisting of "contributing" to the constitution of the extension (of tokens) of a target category a specific categorical subdimension extracted from each precursor category. This contribution depends on the intensity of the attentional focus directed at the precursor category. The categorical segment specific to a target neuron is thus constituted (in terms of its token extension) by assembling semantically complementary categorical subsegments extracted from the extensions of subordinate categories.

Finally, the "$\Sigma$ effect" or synthetic categorical phasing, which refers to the phenomenon of synthetic cognition whereby identical subsegments of precursor categorical segments (i.e., tokens from neurons in layer $n$) that are simultaneously activated in the target neurons (layer $n+1$) enter into categorical resonance. This resonance partially governs the determination of the tokens constituting the extensions of the categories of these target neurons. From a qualitative perspective, this process manifests as a phenomenon of categorical intersection, genetically defining the content (in terms of tokens) of the categorical extension of the target categories, i.e., the categorical segmentation performed by the involved target neurons. The extraction of categorical subdimensions performed here within the precursor categories involves extracting common subdimensions shared by these precursor categories, rather than extracting distinct and complementary subdimensions within each of these precursor categories, as is the case with the "w effect."

These three causal mathematical-cognitive factors of categorical segmentation, as mentioned, orchestrate at the level of a target neuron (layer $n+1$) a mechanism for extracting specific categorical subdimensions from the categories carried by its precursor neurons (layer $n$). These extracted precursor categorical subdimensions, assembled by the aggregation function at the level of the target neuron, thereby define, part by part, the content of the extension (in terms of tokens) of the category of this target neuron, that is, the categorical segment specifically associated with this target neuron. This synthetic conceptual mechanism of concept extraction, extensively studied in cognitive psychology through its human equivalents \cite{Bolognesi2020, Haslam2020, EysenckKean2020, Fel2024, Bathia2024, Marconato2024, Zettersten2024}, is fascinating from an epistemological perspective and contributes to the construction of the "reality" produced by artificial cognition in its interaction with the token world it encounters.

Our preliminary study, which we have just summarized, investigated the process of synthetic categorical extraction by attempting to shed light on how it results from an activity performed simultaneously, at the level of a target neuron, on its respective precursor neurons. In other words, we studied how the extension (of tokens) of a target neuron's category is generated by the joint interaction (in terms of activation, intersection, or categorical complementation) of the extensions of the categories of its precursor neurons, each precursor contributing a part of this extension. In the context of the present, more localized research, we now aim to better understand how this phenomenology of categorical extraction manifests "semantically" in terms of categorical abstraction at the local level of the category of each given precursor neuron; that is, in terms of the extraction of certain tokens (constituting a precise categorical subdimension) and not others, which then form the "categorical background" from which the specifically abstracted subdimension of a given precursor category is detached. The aim is thus to understand how a "categorical outlining" is performed on the relative categorical variability of the tokens constituting the extension of the category of each precursor neuron in order to extract, from each of them, a subset of tokens that are categorically homogeneous (with respect to a given categorical subdimension) and aligned with the (new) specific category uniquely constructed by their corresponding successor neuron.

\section{Categorical Extraction, Conceptualization, and Abstraction}
The capacities for abstraction and conceptualization constitute a fundamental categorical extraction ability of human cognition. They play a central role in intelligence \cite{Konidaris2019} and make reasoning, generalization, and problem-solving possible \cite{Dehaene2022}. They also provide a key skill for learning, robust adaptation of knowledge to a new domain, and analogy \cite{Mitchell2021, Hofstadter2013}.

While human beings demonstrate a strong aptitude for abstraction and conceptualization, composing new concepts from prior ones, the execution of these cognitive processes is more challenging for artificial intelligence systems \cite{Tang2023}. Indeed, whereas humans excel at extracting abstract patterns from different sequences, filtering out irrelevant details, and transferring these generalized concepts to other sequences, artificial neural networks are known to encounter greater difficulties in this area \cite{Wu2024, Xie2024}.

Within the classical theory of concepts \cite{Margolis2019}, philosophy defined abstraction as the extraction of definitions (i.e., defining traits) constituting necessary and sufficient conditions for an element's membership in the extension of a concept. Empiricist philosophers later conceived abstraction as the result of extracting common traits from various sensory experiences or their mnemonic storage, with these abstractions arising from a "distillation" process applied to concrete impressions embedded in perception.

The question of abstraction is, de facto, epistemologically closely linked to that of categorization \cite{Pichat2024b, Pichat2024c}. In the field of cognitive and developmental psychology, various approaches to the nature of the content and organization of concepts are explored, each focusing on specific (complementary or divergent) aspects of conceptual entities: core knowledge \cite{Carey2011, Spelke2007}, exemplars and prototypes as nodal entities calibrating concepts \cite{Nosofsky1986, Rosch1999, Singh2020, Vogel2021, Nosofsky2022}, semantic networks defining the structuring of concepts among themselves \cite{Collins1969, Collins1975, Hornsby2020}, the finalization of categories from a situated cognition perspective \cite{Barsalou2009, Glaser2020, Love2022}, the metaphorized dimension of concepts \cite{Lakoff2008}, concepts as mental models rooted in perceptual activity \cite{Fauconnier2017}, and the relationship of concepts to a language of thought within a probabilistic stochastic approach \cite{Goodman2015}. These various approaches each highlight different modalities, functionalities, or purposes of abstraction in its relationship to categorization, but it is arguably Vergnaud \cite{Vergnaud2009, Vergnaud2016} who connects them most directly by emphasizing that conceptual categorization is fundamentally an activity of abstraction and extraction of characteristics deemed relevant (concepts-in-act) or true (theorems-in-act) within the context of a given activity.

\section{The Nature of Contents Extracted by Abstraction and Conceptualization}
But what does the categorical extraction constitutive of abstraction and conceptualization translate to? In other words, what is the cognitive or categorical nature of what is extracted in the process of abstraction or conceptualization?

In an empiricist-inspired epistemology, the process of abstraction is often positioned within a concrete/abstract dichotomy, with categorical abstraction considered in terms of extracting more ethereal elements from tangible ones. For instance, Cuccio \cite{Cuccio2018}, in the field of neuroscience, suggest that abstract categories are generated based on concrete concepts that constitute previously constructed abstractions. In this vein, Pulvermuller \cite{Pulvermuller2018}, while highlighting the limitations of amodal approaches purely related to definitional traits and disconnected from the context of objects and actions, sheds light on neurobiological mechanisms for extracting common semantic traits, emphasizing the anchoring of the symbolic in entities of the "real" world. This is achieved through a model identifying the processes underlying the rooting of symbols, categories, and concepts in real-world entities, in terms of associations between signs and material reference elements, resulting in the formation of abstract representations through the extraction of semantic traits. The author then points to non-symbolic representations in the form of emerging semantic circuits dispersed across various specific cortical regions, depending on the semantic nature of the sign involved (e.g., action words => motor areas; sensory words => sensory areas).In the specific context of machine learning, studies related to vision and image processing are a rich source of investigations into the nature of categorical extraction. These studies demonstrate the extraction of basic traits (color, texture), complemented by more complex extracted features \cite{Tater2024, Brysbaert2014, Lynott2020, Kastner2020, Harrison2023}; the combination of simple categories (with a simple and identifiable reference, perceived by the senses), when sufficiently numerous, into more abstract, extracted categories \cite{Xie2024}; or the construction and generalization of abstract representations from concrete sequences by extracting and segmenting shared traits of items appearing in the same context \cite{Wu2024}.

Various studies, particularly in the field of neurobiology, describe the process of abstraction as the extraction of cognitive maps of mental spaces related to a given domain of knowledge \cite{Stoewer2022, OKeefe1971, Epstein2017, Killian2018, OKeefe1978, Moser2017}; these maps being, for instance, generated by biological neurons (place and grid cells) that enable the representation of experiences and memories. Different techniques for simulating the emergence of these cognitive maps extracted from data can be implemented using formal neural networks that learn based on statistical units whose characteristics are encoded by feature vectors. Among these techniques, dimensionality reduction and clustering methodologies applied to these vectors are particularly prominent.

In their fascinating study, Ponomarev and Agafonov \cite{Ponomarev2022} (building upon Sousa et al.\cite{Sousa2021}), Ponomarev and Agafonov model the conceptual extraction performed by neural networks in terms of ontologies specifically localized at the level of activations in artificial neural layers. This modeling, in the domain of visual processing, results in maps of compositions of arguments and conceptual predicates extracted from precise neuronal zones within these layers. \cite{Fel2023}, on the other hand, emphasize the importance of local concept studies for a detailed understanding of the nature and organization of conceptual characteristics extracted by artificial networks. By leveraging clustering graphs applied to the activation vector space, they visualize the main features used by a vision model to extract concepts. For instance, as the authors explain, the visual concept of "espresso" is derived from traits such as "bubbles and foam on the coffee," "latte art," "transparent cups with foam and black liquid," "coffee cup handle," and "coffee in the cup." This approach goes beyond classical methods like "heat maps," which focus more on illustrating the "where" rather than the "what" of categorical extraction (\cite{Ghorbani2019}; \cite{Bhatt2020}).

Finally, let us mention structural and formal studies that finely reveal the nature of categorical extractions performed by artificial neurons in terms of conceptual subspaces, constitutive of defining traits, extracted from raw perceptual data \cite{Clark2021}; or, in a similar vein, works modeling conceptual extraction in terms of a representation space endowed with dimensions or separable subspaces constitutive of conceptual traits \cite{Higgins2017, Higgins2018}.

\section{Problematic}
\subsection{Categorical Outlining}

In the field of image or video editing and graphic design, the term outlining refers to the process of separating an element of a visual scene (static or dynamic) from its background. This is done to highlight the isolated element or to manipulate it subsequently, for example, by integrating it into another visual field, emphasizing certain details, altering specific color nuances, or improving its focus. Outlining fundamentally involves an operation of delimiting the contour of what is instantiated as the relevant object to dissociate it from what is defined as background or noise, deemed non-relevant, to use terminology from signal processing.

By analogy, we define, in the domain of synthetic cognition, categorical outlining as the creation, instantiation, and enaction (in Varela's sense)\cite{varela1984} of a singular (sub-)dimension within the infinite-dimensional vector space of characteristics that can be attributed to a given entity. Here, we are indeed referring to original creation and not the identification of a pre-existing (sub-)dimension already available within a pre-given world of characteristics endowed with ontological existence \cite{Varela1988}. Furthermore, it should be noted that this (sub-)dimension may or may not be analogous to a category of thought currently existing in human cognition.

\subsection{Categorical Outlining and the Construction of the\\ Form/Background Separation}

The previously mentioned studies in the fields of human psychology, neuroscience, and neural networks explore with interest the nature of the contents and the organization of elements extracted through abstraction and conceptualization in terms of exemplars and prototypes, mental models, basic traits derived from concrete elements, emergent semantic circuits, cognitive maps of mental spaces, ontologies, representational clusters, and more. These various approaches showcase diverse and fascinating models of the results of these extracted contents. But through what processes are these contents extracted by the formal neurons themselves? How does the process of categorical outlining manifest cognitively or categorically with respect to these contents? Through what phenomenology does synthetic cognition’s "decision" of what constitutes the conceptual form to be retained and dissociated from a background manifest within the framework of its abstraction and conceptualization activity?

In the field of human neuroscience, Savioz et al. \cite{Savioz2010} describe aspects of the effect of the activation function on the construction of the dissociation between background and form, specifying how neuromodulators influence synaptic plasticity \cite{Magee2020}. Indeed, dopamine and noradrenaline induce a strengthening of activated synapses and, conversely, a weakening of non-activated synapses, thereby generating a contrast between the signal and the background noise. This process can be modeled by a sigmoidal transfer function \cite{ServanSchreiber1990}, equipped with a gain parameter $G$. An increase in $G$ leads to stronger neuronal activations in response to positive inputs and stronger inactivations in response to inhibitory inputs. As a result, inputs are better discriminated, and relevant stimuli are more effectively detected against a background noise \cite{Hock2024, Green2024}. This results in more distinct cortical representations and better inhibition of irrelevant information. Zeki \cite{Zeki2002}, for instance, demonstrates the effect of this figure-background extraction process in the cortical processing of visual information at the level of occipital areas V1 to V4.

In the field of human neuroscience, Savioz et al. \cite{Savioz2010} describe aspects of the effect of the activation function on the construction of the dissociation between background and form, specifying how neuromodulators influence synaptic plasticity \cite{Magee2020}. Indeed, dopamine and noradrenaline induce a strengthening of activated synapses and, conversely, a weakening of non-activated synapses, thereby generating a contrast between the signal and the background noise. This process can be modeled by a sigmoidal transfer function \cite{ServanSchreiber1990}, equipped with a gain parameter $G$. An increase in $G$ leads to stronger neuronal activations in response to positive inputs and stronger inactivations in response to inhibitory inputs. As a result, inputs are better discriminated, and relevant stimuli are more effectively detected against a background noise \cite{Hock2024, Green2024}. This results in more distinct cortical representations and better inhibition of irrelevant information. Zeki (2002), for instance, demonstrates the effect of this figure-background extraction process in the cortical processing of visual information at the level of occipital areas V1 to V4.

These elements demonstrate that the outlining process in human cognition fundamentally involves a quantitative mechanism: stronger inputs are amplified, while weaker inputs are inhibited. The same is true in synthetic cognition insofar as it is also driven by an activation function composed with an aggregation function. The aggregation function causally governs, as previously mentioned through the mathematical-cognitive factors of priming, attention, and categorical phasing, the intensity of activation specifically allocated to certain tokens as opposed to others. But how can this quantitative phenomenology be better understood qualitatively? Through what cognitive and categorical qualitative properties does this quantitative process of determining and constructing what is the "form" relevant to extract from a "background"—both cognitively constructed—manifest? In other words, at the level of a precursor neuron, what are the properties (resulting from the aggregation of one of its associated target neurons) of the mechanism of abstraction of a categorical subdimension (the retained categorical form) within the categorical dimension carried by this precursor neuron?

\subsection{Reflective Abstraction and the Cognitive-Mathematical Operative Mode of Categorical Outlining}
Jean Piaget is certainly one of the researchers in the field of human psychology who has most profoundly investigated the concept of abstraction. The theoretical framework he proposes provides a valuable heuristic basis for reflecting on the operative mode of the categorical outlining process in its activity of differentiating a form (categorical dimension) from a background within the domain of synthetic cognition. Piaget \cite{Piaget2001} distinguishes two types of abstraction (we do not address here the case of reflective abstraction nor the sub-case of pseudo-empirical abstraction).

Simple abstraction (also called empirical abstraction) operates on the "material" and "immediately observable" dimensions of a set of objects (or actions). It focuses on physical dimensions "imposed" by perception and "inherent" to the object (such as its weight, texture, or color) or to the action (such as its direction or force). This type of abstraction is partly related to that postulated by philosophical empiricism. However, only partly, because Piaget's constructivism specifies that it nonetheless requires knowledge frameworks that were previously generated by reflective abstraction. For example, color is not a fully immediate datum but presupposes a categorization and serialization of impressions derived from perceived varied wavelengths—a categorization not directly extracted from reality by empirical abstraction \cite{Montangero1994}. Similarly, even in physics, measured quantities (mass, force, acceleration, etc.) are themselves constructed and, therefore, the result of inferences derived from prior reflective abstractions \cite{Piaget1974}.

Reflective abstraction, on the other hand, is the activity of identifying a dimension associated with an object and then utilizing this dimension as an element within a broader structure that differs from the mere framework of perception (unlike simple abstraction). More precisely, reflective abstraction pertains to mental activities and coordinations that an individual performs on extracted dimensions, no longer requiring the proximal support of the objects or actions to which these dimensions are associated \cite{Nisa2020}. Consequently, what is extracted here is not information closely tied to characteristics of the world but rather modes of structuring that the individual has themselves introduced into reality. In this context, Piaget illustrates this with the transition from arithmetic to algebra, which involves an abstraction of operations and numerical relationships without requiring actual numbers to apply them to.

A process of knowledge formation that is more endogenous in nature than empirical abstraction, Piagetian reflective abstraction unfolds in three stages \cite{Montangero1994}: (i) abstraction proper, (ii) reflection, and (iii) reasoning. Abstraction proper does not pertain to a property of the physical world (e.g., color) and is therefore not an empirical abstraction. It involves extracting a property from the individual's own activity, such as coordinating elements (e.g., grouping, ordering, or matching). This initial phase of abstraction produces either new knowledge (through objectification: a mental instrument becomes an object of thought, a new concept) or a new tool of thought (a schema). The second phase, reflection, translates into a projection of what has been extracted at the lower level onto a higher plane of knowledge, distinguished by its nature or complexity. This transposition to a more elaborate plane enables a new, more abstract, and decontextualized utilization of the extracted dimension or element. The final phase, reasoning, involves genuine reconstruction and reorganization of the dimension or element projected onto the higher plane. This dimension or element can then (i) be translated into the (more abstract) terms of the new plane and (ii) be mentally manipulated as elements among others, allowing for their combination with other dimensions or elements also extracted.

\subsection{Categorical Outlining and Synthetic Reflective Abstraction}

The three stages of reflective abstraction, as defined by Piaget, shed light on the qualitative modalities by which the aggregation function $\Sigma(w_{i,j} x_{i,j}) + b$ in artificial neural networks "separates" a form from a background—a singular subdimension from an infinite range of possible subdimensions of categorical segmentation of the world (tokens). Indeed, the activity of this aggregation function can be interpreted as operating these three stages as follows. First, through an initial stage of abstraction ("proper abstraction"), the aggregation function extracts a dimension not inherently present in the tokens themselves but resulting from the activity of the neural network itself—a categorical dimension $x_{i,j}$ constructed within the activation space of each neuron in a layer $n$. Then, in a second stage, the aggregation function, through a process of reflection, projects this extracted categorical dimension onto a new (representational) vectorial plane, more abstract—the superordinate neurons of layer $n+1$. Finally, in a third and final stage of reasoning, and within this new vectorial plane, the extracted dimensions are subjected to further operations of weighting $w_{i,j}$, summation $\Sigma$, and bias addition $b$. These operations, coupled with the activation function $f_a$, create an entirely new dimension $f_a(\Sigma(w_{i,j} x_{i,j}) + b)$. This new dimensional form is specifically constructed at this stage of reasoning (in Piaget's sense) from the categorical subdimension uniquely outlined from each categorical dimension (the categorical background) of the precursor neurons in layer $n$.

As beautifully written by von Glaserfeld \cite{Glaserfeld2002}, citing von Humboldt: "in order to reflect, the mind (...) must grasp as a unit what was just presented, and thus posit it as an object against itself. The mind then compares the units, of which several can be created in that way, and separates and connects them according to its needs" (p. 90). The author elaborates further (pp. 90-91): "'to grasp as a unit what was just presented' is to cut it out of the continuous experimental flow. In the literal sense of the term, this is a kind of abstraction (…). Focused attention picks a chunk of experience, isolates it from what came before and from what follows, and treats it as a closed entity." Thus, categorical outlining operating at the level of the category of each precursor neuron does not "tear out" but rather genuinely fabricates a categorical subdimension on the basis of a reflective abstraction activity. This process manifests as a unique, weighted recombination of input categorical dimensions.

But, again and always, how does this reflective abstraction, underpinning the categorical outlining performed by each neuron in layer $n$ through its specific aggregation function, manifest categorically, cognitively, or even semantically? This is the question to which we will attempt to provide insights, chosen here in terms of semantic reduction, separation of embeddings in the vector spaces of GPT-2 XL, reorganization of these embedding dimensions, and the semantic nature of the categorical dimensions extracted. These are the properties and phenomenologies of categorical outlining vectored by synthetic reflective abstraction that we will endeavor to begin exploring in this investigation.

\section{Methodology}
\subsection{Methodological Positioning}

To methodologically situate our present exploratory study, we provide here a concise overview of various explainability methods aimed, with different levels of cognitive detail, at extracting the content or informational processes of formal neural networks, whether they are structured by layers, in groups, or as a complete network.

Broad-spectrum cognitive research focuses on analyzing the gaps between inputs and outputs, seeking to elucidate the connection between initial data and outcomes in a language model. Among these approaches, gradient-based methods assess the role of each input datum by leveraging derivatives for each input dimension \cite{Enguehard2023}. The characteristics of the inputs can be evaluated through elements such as features \cite{Danilevsky2020}, token importance scores \cite{Enguehard2023}, or attention weights \cite{Barkan2021}. Additionally, example-based approaches examine how outputs evolve in response to different inputs by observing the effects of slight modifications to the inputs \cite{Wang2022} or alterations such as deletion, negation, mixing, or masking of inputs \cite{Atanasova2020, Wu2020, Treviso2023}. Some studies also focus on the conceptual mapping of inputs to assess their contribution to the observed outputs \cite{Captum2022}.

Methods with finer cognitive granularity focus on the intermediate states of the language model rather than its final output, analyzing the partial outputs or internal states of neurons or groups of neurons. In this context, some approaches analyze and linearly decompose the activation score of a neuron in a specific layer in relation to its inputs (neurons, attention heads, or tokens) in the previous layer \cite{Voita2021}. Others aim to simplify activation functions to improve their interpretability \cite{Wang2022}. Certain techniques, leveraging the model's vocabulary, focus on extracting encoded knowledge by projecting intermediate connections and representations through a correspondence matrix \cite{Dar2023, Geva2023}. Finally, some methodologies rely on statistics of neuronal activation in response to datasets \cite{Bills2023, Mousi2023, Durrani2022, Wang2022, Dai2022}. Our current exploratory study specifically falls within this latter group of approaches.

\subsection{Methodological Option}

In our exploratory research, we chose to investigate the GPT model developed by OpenAI, focusing specifically on the GPT-2XL version. This choice is justified by the fact that GPT-2XL offers sufficient complexity to analyze advanced synthetic cognitive phenomena while remaining less complex than GPT-4 or its current multimodal version, GPT-4o. A practical aspect also influenced our decision: in 2023, OpenAI made detailed information available, as highlighted in the work of Bills et al. \cite{Bills2023}, regarding the parameters and activation values of the model's neurons, which are essential data for our study.

To simplify our exploration, we focused on examining the first two layers of GPT-2XL, each containing 6400 neurons, for a total of 12800 artificial neurons. Regarding the tokens and their activation values among these neurons, we chose to analyze, for each neuron, the 100 tokens with the highest average activation values, which we referred to as "core-tokens."

To study the semantic proximity between tokens in the context of the preliminary static results we will present below, we made the central choice to measure the cosine similarity within the GPT-2XL embedding base, rather than in the base of GPT-4, which is more powerful, in order to avoid falling into the methodological limitation mentioned by Bills et al. \cite{Bills2023} and Bricken \cite{Bricken2023}, which involves matching synthetic cognitive systems that do not rely on the same embedding system, that is, not on the same categorical segmentation system. However, for comparison and verification of the plausibility of our data, we also used two other classic, freely available embedding bases: Alibaba-NLP/gte-large-en-v1.5 and the BERT base model.

\subsection{Statistical Choices}

For our statistical analyses, we used Python libraries from the SciPy suite, following the recommendations of Howell \cite{Howell2024} and Beaufils \cite{Beaufils1996}.

To assess the normality of our data, a prerequisite for conducting parametric tests, we adopted a two-pronged approach. First, we conducted inferential tests: the Shapiro-Wilk test, suitable for small samples; the Lilliefors test, useful when the parameters of the normal distribution are unknown and estimated from the data; the Kolmogorov-Smirnov test, ideal for large samples; and the Jarque-Bera test, which assesses the symmetry and kurtosis of the data for large samples. Second, we supplemented this analysis with descriptive indicators such as skewness and kurtosis, and graphical methods like the QQ-plot to compare the observed distributions to a theoretical normal distribution. Regarding the evaluation of homoscedasticity (equality of variances between groups), we used Bartlett’s test (sensitive to deviations from normality) complemented by Levene’s test (less sensitive to non-normality).

The results, partially indicated in the following sections of this article, reveal a partial normality of our data. As a result, we primarily employed the following for our group comparisons and distribution studies:
\begin{itemize}
    \item The Kruskal-Wallis test, which examines the relationship between a nominal variable defining $k$ independent groups and a ranking variable. This was applied by ranking our numerical activation data for the tokens, while ensuring the condition of group sizes strictly greater than 5.
    \item The univariate $\chi^2$ goodness-of-fit test, applied while respecting its conditions regarding theoretical and observed frequencies, thus not requiring adjustments for small sample sizes (Fisher’s or Monte Carlo statistics).
\end{itemize}

Regarding our statistical study of the preferential orientation of the embedding dimension vectors of the tokens based on their characteristic of being "taken-tokens" or "left-tokens," we employed a dimensionality reduction approach using principal component analysis (PCA). A PCA was performed for pairs (precursor neuron – target neuron); the precursor neurons here are neurons from the layer, and the target neurons are the 10 neurons from layer 1 with which each precursor neuron has the highest connection weight. For each PCA, the statistical units involved are the 100 core-tokens of the precursor neuron, i.e., the 100 tokens with the highest average activation within this neuron. The variables used here are the 1600 embedding dimensions of GPT-2XL, supplemented by two antagonistic dichotomous variables (0/1): "taken-token" and "left-token." A core-token of a precursor neuron is a taken-token if it is also a core-token of its associated target neuron, or a left-token if it is not. The usual conditions for applying PCA are: (i) a number of statistical units greater than 100, (ii) a number of statistical units greater than 10 times the number of variables involved, (iii) sphericity confirmed by the significance ($\alpha = 5\%$) of Bartlett’s test, and (iv) a global adequacy validated by a Kaiser-Meyer-Olkin (KMO) coefficient greater than .5 or even .7. These conditions were only partially met in the context of our current study, so the results should be considered with caution and viewed as heuristic for considering complementary studies, such as the complementary t-STN study, which will be directly presented in the next section of this article.

Our PCA parameterization options were as follows: (i) orthogonal (varimax) and oblique rotation tests, (ii) prior data reduction to normalize them, (iii) verification of the quality of variable representation on the factors (communalities cos2) greater than 40\%, to avoid excessive variance loss during the vectorial projection onto the factors, (iv) examination of correlations that are too strong ($|\rho| > .9$), to eliminate potentially redundant variables, (v) the Kaiser criterion ($\lambda > 1$) combined with the rule for the percentage of variance explained ($\sigma^2 > .6$) to determine the number of factors to retain, (vi) over-weighting of the two variables "taken-token" and "left-token" by 1 \% (of the number of embedding dimensions), to format the PCA so it produces the desired factor axis $F_2$ related to whether a token is a "taken-token" or "left-token", and (vii) selection of the 1671 precursor neurons in layer 0 with a percentage of taken-tokens between 15\% and 85\%, to ensure a minimal number of taken-tokens present.

To further explore how the 1600 dimensions of the token embeddings tend to be preferentially distributed in the vector space of GPT-2XL embeddings based on whether these tokens are taken-tokens or left-tokens, and considering the methodological limitations previously mentioned regarding our PCA study, we employed another dimensionality reduction approach, the t-SNE technique. A significant advantage of this method is that it does not presuppose a linear combination of the original variables in the construction of the factor axes, and is significantly less restrictive in terms of application conditions. The statistical units and variables here are the same as those mentioned for our PCA study, but without the inclusion of the two additional variables "taken-token" and "left-token". Similar to the previous technique, a t-SNE study was performed on the 1671 precursor neurons mentioned earlier.

\subsection{Objective and Implementation of the Study in Terms of Statistical Observables}

A categorical dimension (in the sense of Varela), from which a singular categorical subdimension will be outlined, can be minimally defined by three types of elements:
\begin{itemize}
    \item Its membership function in the sense of fuzzy logic \cite{Zadeh1996, Wu2022}, defined as follows. Let $X$ be a set, and $A$ a fuzzy subset of $X$ characterized by a membership function $\mu_A : X \to [0, 1]$, determining the partial membership level (as opposed to classical set theory) of an element $x$ in $X$ to the set $A$. This fuzzy set is endowed with a height $h(A) = \max\{\mu_A(x) \mid x \in X\}$ and a kernel $\text{noy}(A) = \{x \in X \mid \mu_A(x) = 1\}$ (if $A$ is normalized, i.e., $h(A) = 1$), containing the elements $x$ fully belonging to $A$. This membership function, in the context of synthetic categorical cognition, is vectorized by the aggregation function (composed with the activation function) associated with the neuron carrying the involved categorical dimension.
    \item Its extension \cite{Nadeau1999}, defined as the set of tokens for which the involved neuron is activated, i.e., the support $\text{sup}(A) = \{x \in X \mid \mu_A(x) > 0\}$. More specifically, the extension of core-tokens, i.e., strongly activated tokens, determined by $\alpha$-cut $(A) = \{x \in X \mid \mu_A(x) \geq \alpha\}$, with $\alpha$ sufficiently high.
    \item Its interpretation, a "meaning" attributed to this dimension via an inferential interpretative construction.
\end{itemize}
\section{Results}
\subsection{Normality of the Data}

In relation to our first batch of statistical studies presented below, concerning categorical reduction operationalized in terms of semantic distances measured from cosine similarity, we performed normality and homogeneity of variance checks on our data. For each target neuron in layer 1, the tests were initiated among its 10 precursor neurons (from layer 0) with high connection weights, considering only the precursors for which both the number of taken-tokens and left-tokens is greater than or equal to 6; this was done to ensure a minimum number of tokens for each category. This results in a theoretical maximum of 6,400 target neurons (in layer 1) x 10 precursor neurons (in layer 0) = 64,000 studies of taken-token clusters (referred to as "taken-clusters"); and an actual number of 9,007 taken-clusters studied (as well as left-clusters for the homoscedasticity case). For each study, the variable involved is the cosine similarity between all the (distinct) tokens within the cluster in question.
Table 1 shows a contrasted normality of cosine similarity between the tokens of different taken-clusters, with compatibility percentages of inferential tests with a normality hypothesis ranging from 52\% to 93\% using the GPT-2XL embeddings, which were the most reliable, and lower results (from 21\% to 73\%) for the other embeddings. Table 2 similarly presents a mixed homogeneity of variances of the tokens between the taken and left-clusters, with 58\% of inferential tests compatible with this hypothesis for GPT embeddings, and lower values for the others (ranging from 29\% to 43\%). These results suggest that we should not rely on parametric tests (except as an indication), as their application conditions are only partially validated.

\begin{table}[H]
    \centering
    \renewcommand{\arraystretch}{1}
    \small
    \rowcolors{1}{gray!20}{gray!40}
    \begin{tabular}{|c|c|c|c|}
        \hline
        & GPT2-XL & Alibaba & BERT \\
        \hline
        \% of $p_{sw} > .05$ & 52.52\% & 21.37\% & 29.57\% \\
        \hline
        \% of $p_{Lilliefors} > .05$ & 64.39\% & 30.48\% & 32.71\% \\
        \hline
        \% of $p_{ks} > .05$ & 92.78\% & 65.36\% & 66.13\% \\
        \hline
        \% of $p_B > .05$ & 79.52\% & 52.60\% & 72.58\% \\
        \hline
        Mean & .537 & .681 & .907 \\
        \hline
        Median & .524 & .661 & .913 \\
        \hline
    \end{tabular}
    \captionsetup{justification=centering, format=plain}
    \caption*{\textit{Table n$^{\circ}$1: Normality ratio statistics of cosine similarities between taken-tokens of each taken-cluster (Layer 1 ; $N=9007$).}}
    \label{tab:table1}
\end{table}

\begin{table}[H]
    \centering
    \renewcommand{\arraystretch}{1.3}
    \rowcolors{1}{gray!20}{gray!40}
    \begin{tabular}{|c|c|c|c|}
        \hline
        & GPT2-XL & Alibaba & BERT \\
        \hline
        \% of $p_{Levene} > .05$ & 58.61\% & 38.86\% & 42.61\% \\
        \hline
        \% of $p_{Bartlett} > .05$ & 58.84\% & 29.12\% & 33.25\% \\
        \hline
    \end{tabular}
    \captionsetup{justification=centering, format=plain}
    \caption*{\textit{Table n$^{\circ}$2: Homoscedasticity ratio statistics of cosine similarities between taken-tokens and left-tokens (Layer 1 ; $N=9007$).}}
    \label{tab:table2}
\end{table}

\subsection{Outlining and Categorical Reduction}
Within the artificial neurons of language models, as we have indicated previously, the abstraction process generated, among other factors, by the aggregation function $\Sigma(w_{i,j} x_{i,j}) + b$, "detaches" a constructed categorical form from a background, a singular categorical subdimension from an infinite range of possible categorical segmentation subdimensions of the world (tokens). This process is generated by the three mechanisms of synthetic cognition, mathematically supported by this aggregation function: priming, attention, and categorical phasing. More specifically, we postulate that the ultimate reflection step of Piagetian reflective abstraction, performed by the aggregation function of a neuron in layer $n+1$, will cause the extraction, the outlining of a particular subdimension from each of its contributing precursor neurons (i.e., sources of taken-tokens) in layer $n$. The successive application, at the level of the target neuron, of these different categorical subsegments thus forms "part by part" the extension of the new category specific to this target neuron, i.e., the set of its specific core-tokens.

This extraction, by a target neuron, of specific categorical subdimensions from the relative categorical diversity of the core-tokens of each of its precursor neurons, should result in the segmentation of taken-clusters that are categorically more homogeneous; at least, more homogeneous in terms of a categorical segment, the specific subdimension extracted from the categorical dimension of each precursor. In other words, we postulate a first characteristic of categorical outlining: categorical reduction, which manifests as the partitioning, within the extension (of tokens) of the category associated with each precursor neuron, of a token cluster (i.e., a taken-cluster) that exhibits less categorical variability compared to the initial dispersion, with this latter being conceptually "narrowed" around the subdimension thus extracted.

We operationalize this postulate through two methodologically similar hypotheses, each providing its own added value. First, for each precursor neuron (contributing) of each target neuron, the involved taken-cluster should present a higher average cosine similarity (measuring the average proximity between all the tokens constituting this cluster) compared to the average cosine similarity of the core-tokens associated with this precursor neuron. Second, for each precursor neuron (contributing) of each target neuron, the involved taken-cluster should present a higher average cosine similarity (measuring the average proximity between all the tokens constituting this cluster) compared to the average cosine similarity of the left-tokens associated with this precursor neuron. We conduct 9007 statistical tests of these two hypotheses, corresponding to the 9007 cases of taken-clusters and left-clusters containing 6 or more tokens (in accordance with the application conditions of the non-parametric Kruskal-Wallis test).

Regarding our first hypothesis, Table 3 (and its visualization in Graph 1) demonstrates a mean superiority of the categorical homogeneity of the taken-tokens compared to that of the core-tokens, with a difference of .14 (calculated using GPT-2XL embeddings), which is relatively significant given that cosine similarity ranges from 0 to 1 in absolute value. A percentage of 95\% of cluster cases show a positive difference $d$ (where $d = \text{mean}(COS(\text{taken-tokens})) - \text{mean}(COS(\text{core-tokens}))$), and this difference is largely significant ($p(\chi^2) < .0001$). Similarly, the percentage of cases showing a significant difference ($p < .05$) in mean cosine proximity is notable (81\% with a Kruskal-Wallis test, and even 86\% with a Student's t-test, although the latter indicator is less relevant as mentioned earlier). Other embedding systems show similar results, though weaker, presumably due to their lower capacity to perform semantic analyses suited to the categorical segmentation mode of GPT-2. These results are compatible with our first hypothesis.

\begin{table}[H]
    \centering
    \renewcommand{\arraystretch}{1}
    \small
    \rowcolors{1}{gray!20}{gray!40}
    \begin{tabular}{|c|c|c|c|}
        \hline
        & GPT2-XL & Alibaba & BERT \\
        \hline
        Mean(Mean(COS(taken-tokens))) & .537 & .681 & .907 \\
        \hline
        Mean(Mean(COS(core-tokens))) & .399 & .585 & .892 \\
        \hline
        Mean(d) & .139 & .097 & .015 \\
        \hline
        \% of $p(t)<.05$ & 85.77\% & 84.13\% & 49.38\% \\
        \hline
        \% of $p(KW)<.05$ & 81.33\% & 78.04\% & 60.73\% \\
        \hline
        \% of $d>0$ & 94.64\% & 93.07\% & 68.05\% \\
        \hline
        $p(\chi^2)$ of $d>0$ & 4.36E-19 & 7.03E-18 & 3.69E-04 \\
        \hline
    \end{tabular}
    \captionsetup{justification=centering, format=plain}
    \caption*{\textit{Table n$^{\circ}$3: Inferential comparison of mean cosine similarity distances between taken-tokens and core-tokens, for each relevant precursor neuron of each destination neuron (Layer 1 ; $N=9007$).}}
    \label{tab:table3}
\end{table}

\begin{figure}[H]
    \centering
    \includegraphics[width=0.9\textwidth]{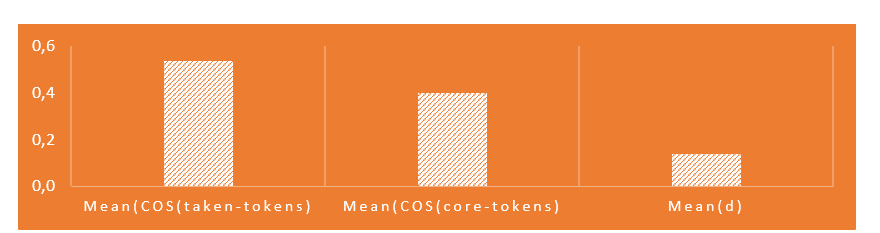}
    \captionsetup{justification=centering, font=small}
    \caption*{\textit{Graph n$^{\circ}$1: Descriptive comparison of mean cosine similarity distances between taken-tokens and core-tokens, for each relevant precursor neuron of each destination neuron (Layer 1 ; $N=9007$).}}
    \label{fig:graph1}
\end{figure}

Table 4 and its associated Graph 2 illustrate these overall results with the case of a precursor neuron of target neuron number 3000 in layer 1.

\begin{table}[H]
    \centering
    \renewcommand{\arraystretch}{1}
    \small
    \rowcolors{1}{gray!20}{gray!40}
    \begin{tabular}{|c|c|c|c|}
        \hline
        & GPT2-XL & Alibaba & BERT \\
        \hline
        Mean(COS(taken-tokens)) & .604 & .678 & .855 \\
        \hline
        Mean(COS(core-tokens)) & .444 & .593 & .889 \\
        \hline
        d & .161 & .085 & -.034 \\
        \hline
        $p(t)$ & 5.41E-06 & 4.26E-06 & 9.87E-01 \\
        \hline
        $p(KW)$ & 3.24E-06 & 2.69E-06 & 2.13E-01 \\
        \hline
    \end{tabular}
    \captionsetup{justification=centering, format=plain}
    \caption*{\textit{Table n$^{\circ}$4: Inferential comparison of mean cosine similarity distances between taken-tokens and core-tokens, for one precursor neuron of neuron 3000 (Layer 1).}}
    \label{tab:table4}
\end{table}

\begin{figure}[H]
    \centering
    \includegraphics[width=0.9\textwidth]{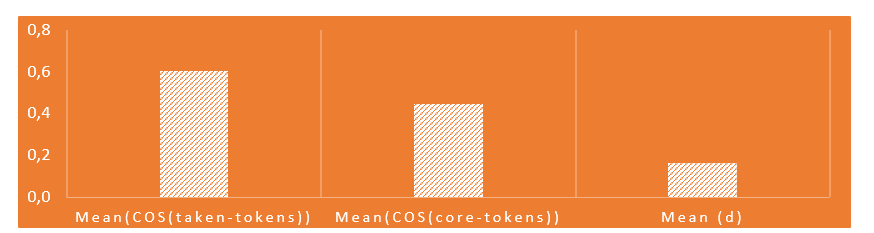}
    \captionsetup{justification=centering, font=small}
    \caption*{\textit{Graph n$^{\circ}$2: Descriptive comparison of mean cosine similarity distances between taken-tokens and core-tokens, for one precursor neuron of neuron 3000 (Layer 1).}}
    \label{fig:graph2}
\end{figure}

Regarding our second hypothesis, Table 5 (and its visualization in Graph 3) shows a higher average value of categorical proximity for the taken-tokens compared to that of the left-tokens, with a difference of .14 again (GPT-2XL embeddings). A significant percentage ($p(\chi^2) < .0001$) of 94\% of cases show a positive $d$. Again, the percentage of cases associated with a significant difference ($p < .05$) in mean cosine similarity is notable (82\% with Kruskal-Wallis and 85\% with Student's t-test). Other embedding systems show similar results, although weaker, presumably due to their lower capacity to perform semantic analyses suited to the categorical segmentation mode of GPT-2. These results, directly methodologically complementary to the first, are necessarily also compatible with our second hypothesis.

\begin{table}[H]
    \centering
    \renewcommand{\arraystretch}{1}
    \small
    \rowcolors{1}{gray!20}{gray!40}
    \begin{tabular}{|c|c|c|c|}
        \hline
        & GPT2-XL & Alibaba & BERT \\
        \hline
        Mean(Mean(COS(taken-tokens))) & .537 & .681 & .907 \\
        \hline
        Mean(Mean(COS(core-tokens))) & .394 & .581 & .892 \\
        \hline
        Mean(d) & .143 & .100 & .015 \\
        \hline
        \% of $p(t)<.05$ & 85.48\% & 83.87\% & 50.11\% \\
        \hline
        \% of $p(KW)<.05$ & 81.65\% & 78.93\% & 63.44\% \\
        \hline
        \% of $d>0$ & 94.02\% & 92.54\% & 67.07\% \\
        \hline
        $p(\chi^2)$ of $d>0$ & 1.33E-18 & 1.77E-147 & 6.00E-04 \\
        \hline
    \end{tabular}
    \captionsetup{justification=centering, format=plain}
    \caption*{\textit{Table n$^{\circ}$5: Inferential comparison of mean cosine similarity distances between taken-tokens and left-tokens, for each relevant precursor neuron of each destination neuron (Layer 1 ; $N=9007$).}}
    \label{tab:table5}
\end{table}

\begin{figure}[H]
    \centering
    \includegraphics[width=0.9\textwidth]{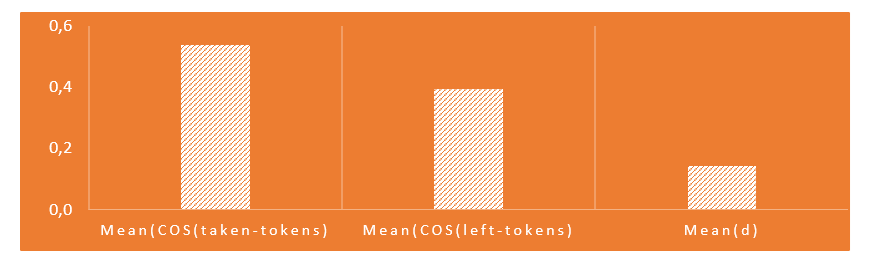}
    \captionsetup{justification=centering, font=small}
    \caption*{\textit{Graph n$^{\circ}$3: Descriptive comparison of mean cosine similarity distances between taken-tokens and left-tokens, for each relevant precursor neuron of each destination neuron (Layer 1 ; $N=9007$).}}
    \label{fig:graph3}
\end{figure}

Table 6 and its corresponding Graph 4 illustrate these overall results, again with the case of a precursor neuron of target neuron number 3000 in layer 1.

\begin{table}[H]
    \centering
    \renewcommand{\arraystretch}{1}
    \small
    \rowcolors{1}{gray!20}{gray!40}
    \begin{tabular}{|c|c|c|c|}
        \hline
        & GPT2-XL & Alibaba & BERT \\
        \hline
        Mean(COS(taken-tokens)) & .604 & .678 & .855 \\
        \hline
        Mean(COS(left-tokens)) & .434 & .590 & .891 \\
        \hline
        Mean(d) & .171 & .088 & -.036 \\
        \hline
        $p(t)$ & 1.34E-06 & 2.10E-06 & 9.94E-01 \\
        \hline
        $p(KW)$ & 9.55E-07 & 1.47E-06 & 1.81E-01 \\
        \hline
    \end{tabular}
    \captionsetup{justification=centering, format=plain}
    \caption*{\textit{Table n$^{\circ}$6: Inferential comparison of mean cosine similarity distances between taken-tokens and left-tokens, for one precursor neuron (Layer 1 ; Control neuron 3000).}}
    \label{tab:table6}
\end{table}

\begin{figure}[H]
    \centering
    \includegraphics[width=0.9\textwidth]{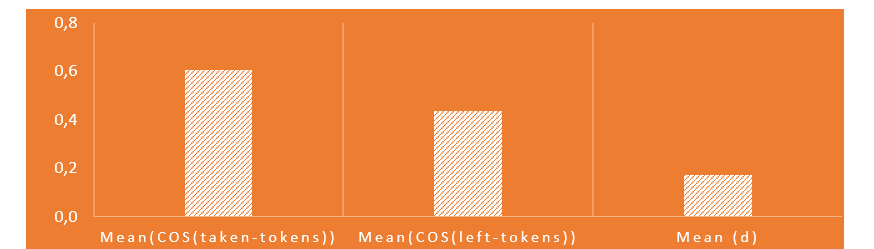}
    \captionsetup{justification=centering, font=small}
    \caption*{\textit{Graph n$^{\circ}$4: Descriptive comparison of mean cosine similarity distances between taken-tokens and left-tokens, for one precursor neuron (Layer 1 ; Control neuron 3000).}}
    \label{fig:graph4}
\end{figure}

The results obtained here are thus compatible with our postulate that the categorical outlining performed by the abstraction carried by the aggregation function involves a categorical reduction process. Specifically, for each target neuron in layer $n+1$, a particular categorical subdimension is extracted from the categorical dimension associated with each of its contributing precursor neurons in layer $n$. This subsegment is characterized by a set of tokens (the taken-tokens) that are more homogeneous among themselves, thereby constituting a specific categorical subdimension rather than a categorically heterogeneous subgroup of tokens or a subgroup of tokens categorically equivalent to the entire set of their core-tokens.

\subsection{Outlining and Categorical Selectivity}

The characteristic of categorical reduction pointed out in the previous section leads us to formulate another postulate regarding the property of categorical outlining carried out by the aggregation function and its abstraction activity, namely, that of categorical selectivity. This categorical selectivity manifests as the extraction, from the category specific to a given precursor neuron, of a subdimension that is very restricted in terms of its own extension, since this precursor category already inherently exhibits a significant categorical convergence that constitutes the categorical dimension it vectorizes.

We operationalize this postulate through the following hypothesis: the cardinality of the token subsets that constitute the taken-clusters tends to be very small compared to the number (100) of core-tokens of each precursor neuron. To test this hypothesis, we set a very low threshold, equal to 6 tokens, which is the reference number against which we will compare the size of each possible taken-cluster associated with layer 1. The total number of taken-clusters involved here is 64,000 (6,400 target neurons in layer 1 x 10 precursor neurons in layer 0).

The table n°7 shows a strong over-representation (86\%) of taken-clusters composed of fewer than 6 taken-tokens, a trend that is highly significant ($p(\chi^2) < .0001$). It should be noted that the very large size of the statistical units involved (64,000) is likely to produce an increased biased significance. However, this bias is not significant as the large effect size at play here can a priori only be associated with high significance. This result is consistent with our hypothesis of categorical selectivity associated with the outlining process.

\begin{table}[H]
    \centering
    \renewcommand{\arraystretch}{1}
    \small
    \rowcolors{1}{gray!20}{gray!40}
    \begin{tabular}{|c|c|c|}
        \hline
        & $<6$ & $\geq 6$ \\
        \hline
        Observed Frequencies & 54993 & 9007 \\
        \hline
        \% of Observed Frequencies & .86 & .14 \\
        \hline
        Expected Frequencies & 3200 & 60800 \\
        \hline
        Weighted $\chi^2$ Residuals & 16.19 & -0.85 \\
        \hline
        $\chi^2$ & \multicolumn{2}{c|}{882406.20} \\
        \hline
        $p(\chi^2)$ & \multicolumn{2}{c|}{.0000} \\
        \hline
    \end{tabular}
    \captionsetup{justification=centering, format=plain}
    \caption*{\textit{Table n$^{\circ}$7: Distribution of taken-clusters with a size lesser than 6 (Layer 1 ; $N=64,000$).}}
    \label{tab:chi2_distribution}
\end{table}

\subsection{Outlining and Separation of Initial Embedding Dimensions}
Let us continue our exploration of the properties of categorical outlining generated by the factors of priming, attention, and categorical phasing carried by the neural aggregation function. The categorical outlining process results ipso facto, at the level of the extension (core-tokens) of the synthetic category associated with each precursor neuron in layer $n$, through the selection and extraction of certain specific tokens (the taken-tokens), which will become part of the extension (core-tokens) of the new synthetic category vectorized by a target neuron in layer $n+1$. In the context of synthetic cognition, this abstraction is certainly not the result of a cognitive processing of the tokens per se (i.e., tokens as a semantic unit, as a human might perceive it), but rather of their coordinates within the embedding dimensions of the input vector space of the involved layer. What is therefore calculated and extracted in a principled and proper sense from the category of a precursor neuron is not the tokens per se but a categorical subdimension within this original categorical dimension.

It is possible to devise a variety of operationalizations to study the characteristics specific to this extracted categorical subdimension (the "categorical form") in relation to those of the non-extracted categorical "residue" (the "categorical background") within the framework of the outlining process. The methodological approach we employ here is based on the analytical framework of the input vector space embeddings of GPT2-XL. This choice can be justified, particularly because we operate at the level of the genetic interaction between layers 0 and 1, which are not yet too "distant" categorically from the categorical dimensions inherent to this initial vector space. Consequently, we will examine, for the 100 core-tokens that constitute the extension of the category of a given precursor neuron in layer 0, the respective characteristics, in terms of these embeddings, of the tokens that are "passed on" or "taken up" by a target neuron (with a strong connection weight) in layer 1. More precisely, we will study the respective characteristics of the tokens, relative to this initial vector space, that become taken-tokens (i.e., those constituting the extracted categorical subdimension from the initial category and subsequently becoming core-tokens of the target category), as opposed to tokens that are not retained (i.e., the left-tokens, which remain part of the non-extracted categorical background).

From a statistical perspective, a relevant approach here is to employ a factorial analysis such as principal component analysis (PCA), with the (1600) embedding dimensions of GPT2-XL taken as variables and the tokens involved as statistical units. This analysis is supplemented by two dichotomous variables: one indicating whether a token is a taken-token and the other indicating whether it is a left-token. We overweigh these two variables (at 1\% of the 1600 embedding variables) to guide the PCA toward producing a factorial axis (with a sufficient eigenvalue) related to whether a token is a taken-token versus a left-token. Subsequently, we analyze how the initial embedding variables distribute with respect to this factor differentiating taken-tokens from left-tokens, exercising caution in our interpretations. This caution is necessary given that, as previously mentioned in the methodology section, the usual conditions for applying PCA are not fully satisfied in our current study. However, within this framework, we retain only those embedding vectors with a representation quality of cos² > .6.

A first application of our approach (precursor neuron no. 12 in layer 0, in its genetic interaction with target neuron no. 816 in layer 1) yields graph no. 5. This correlation circle is highly instructive. It reveals a clear differentiation between a first horizontal factor (explaining 63\% of the variance) clustering embedding vectors (with good representation quality) and a second vertical factor (accounting for 17\% of the extracted variance) opposing the taken-tokens (in green) to the left-tokens (in red). The embedding vectors tend to distribute preferentially based on the "taken" or "left" nature of the tokens involved, producing a graph of dimension extraction from the embedding vector space: taken-tokens are more correlated with certain embedding dimensions, while left-tokens are more correlated with others. It should be noted that these correlations are understandably not very strong, given that (i) the vector space of categorical dimensions generated by layer 1 begins to diverge categorically from the initial embedding vector space (otherwise, the former would add no value over the latter), and (ii) the nonlinear functions also involved (GELU) necessarily distort the original vector topology.

\begin{figure}[H]
    \centering
    \includegraphics[width=0.9\textwidth]{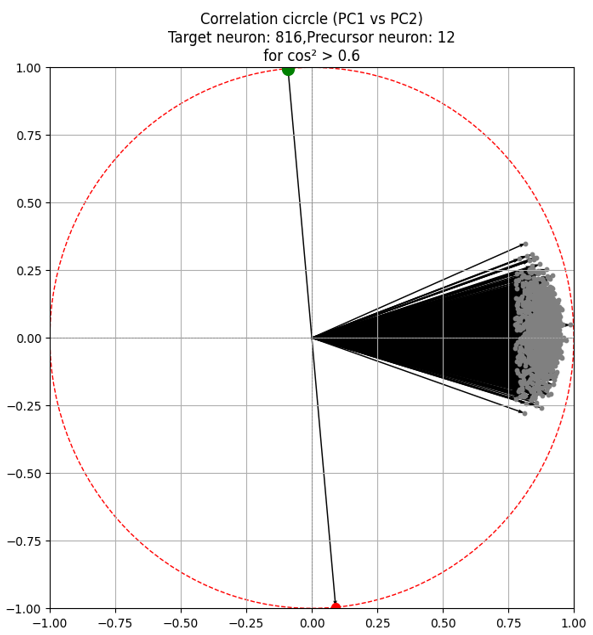}
    \captionsetup{justification=centering, font=small}
    \caption*{\textit{Graph n$^{\circ}$5: PCA correlation circle on embedding variables and group membership variable (left/taken) associated with left-tokens and taken-tokens (Weighting=1\% ; Target neuron 816 ; Precursor neuron 12 ; on the 78.91\% of variables with cos$^{2} > .6$).}}
    \label{fig:graph5}
\end{figure}

Table no. 8, associated with the previous graph, highlights a minority of embedding variables associated with taken-tokens (26\%) compared to left-tokens (74\%). This observation aligns, albeit in a different context, with our earlier postulate of categorical selectivity in outlining. Following the insights from the graph, the averages of dot products reveal a projection that, while relatively weak (for the reasons mentioned above), remains significant for embedding variables onto the taken vs. left-token opposition variable. Consistently with our postulate of categorical selectivity, the mean dot products are lower for taken-tokens than for left-tokens.

\begin{table}[H]
    \centering
    \renewcommand{\arraystretch}{1}
    \small
    \rowcolors{1}{gray!20}{gray!40}
    \begin{tabular}{|c|c|c|}
        \hline
        & Taken-tokens & Left-tokens \\
        \hline
        \% of projected variables & 26.13 & 73.87 \\
        \hline
        Mean(cos) & .0622 & .1015 \\
        \hline
        Mean(scalar product) & .0628 & .1025 \\
        \hline
    \end{tabular}
    \captionsetup{justification=centering, format=plain}
    \caption*{\textit{Table n$^{\circ}$8: Projection statistics of PCA coordinates of embedding variables on group membership variable (left/taken) (Weighting=1\% ; Target neuron 816 ; Precursor neuron 12 ; on the 78.91\% of variables with cos$>.6$).}}
    \label{tab:table8}
\end{table}

Purely descriptive and related to a single neuron pair, the data obtained above tend to be compatible with the following idea: if the outlining process is observed from the reference frame constituted by the initial embedding vector space of GPT2-XL, this process appears to manifest as follows. The abstraction, by a target neuron, of a categorical subdimension (associated with taken-tokens) from the categorical dimension carried by a given precursor neuron correlates with the selective extraction of certain (minority) embedding dimensions that are categorically specifically linked to this abstracted subdimension. Another example (precursor neuron no. 14 in layer 0, in its genetic interaction with target neuron no. 3887 in layer 1) produces similar data and interpretation (cf. graph no. 6 and table no. 9).

\begin{figure}[H]
    \centering
    \includegraphics[width=0.9\textwidth]{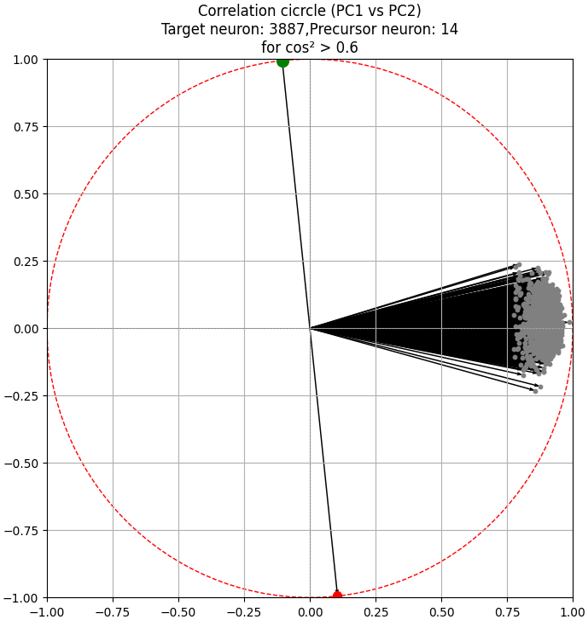}
    \captionsetup{justification=centering, font=small}
    \caption*{\textit{Graph n$^{\circ}$6: PCA correlation circle on embedding variables and group membership variable (left/taken) associated with left-tokens and taken-tokens (Weighting=1\% ; Target neuron 3887 ; Precursor neuron 14 ; on the 86.61\% of variables with cos$^{2} > .6$).}}
    \label{fig:graph6}
\end{figure}

\begin{table}[H]
    \centering
    \renewcommand{\arraystretch}{1}
    \small
    \rowcolors{1}{gray!20}{gray!40}
    \begin{tabular}{|c|c|c|}
        \hline
        & Taken-tokens & Left-tokens \\
        \hline
        \% of projected variables & 14.74 & 85.26 \\
        \hline
        Mean(cos) & .0338 & .0888 \\
        \hline
        Mean(scalar product) & .0342 & .0896 \\
        \hline
    \end{tabular}
    \captionsetup{justification=centering, format=plain}
    \caption*{\textit{Table n$^{\circ}$9: Projection statistics of PCA coordinates of embedding variables on group membership axis (left/taken) (Weighting=1\% ; Target neuron 3887 ; Precursor neuron 14 ; on the 86.61\% of variables with cos$>.6$).}}
    \label{tab:table9}
\end{table}

Let us now examine whether these local trends, observed at the level of certain neurons taken as examples, persist at a more global level. To this end, we applied our PCA approach to the entirety, where applicable, of genetic pairs (core-tokens of the category of a precursor neuron in layer 0 and the "taken" or "left" status of these tokens relative to an associated target neuron (with a strong connection weight) in layer 1). For statistical reasons, not all combinations are feasible, and we retained only cases where (i) 15\% to 85\% of the initial core-tokens became taken-tokens ($N=1671$), and (ii) the KMO index was greater than .5 ($N=950$). It should be noted that, among these 950 cases, only 22\% of the core-tokens involved in layer 0 become taken-tokens (and thus core-tokens) in layer 1, which is again fully consistent with our earlier postulate of categorical selectivity.

Graph no. 7, following the compilation of all obtained data, rotation (with the horizontal axis becoming the one opposing taken and left-tokens), and reconstruction of an average vector of embedding dimensions associated with taken-tokens (with a relatively low norm of .45) and an average vector of embedding dimensions associated with left-tokens (with a norm of .5), confirms our principal trend. On the mentioned global scale, we observe again that the partition of taken-tokens versus left-tokens is associated with a segmentation of the initial embedding vectors of GPT2-XL. In other words, the categorical outlining performed by the aggregation function, which consists of extracting a categorical subdimension (whose extension corresponds to the cluster of taken-tokens specifically involved in each case), results in this singularly extracted subdimension being associated with specific embedding dimensions, selectively abstracted from the entirety of possible initial embedding dimensions. However, in alignment with Piaget's ultimate stage of "reflection" in reflective abstraction, we must bear in mind that the outlined categorical subdimensions are not merely elective extractions but rather authentic original recombinations of these dimensions via the aggregation function.

\begin{figure}[H]
    \centering
    \includegraphics[width=0.9\textwidth]{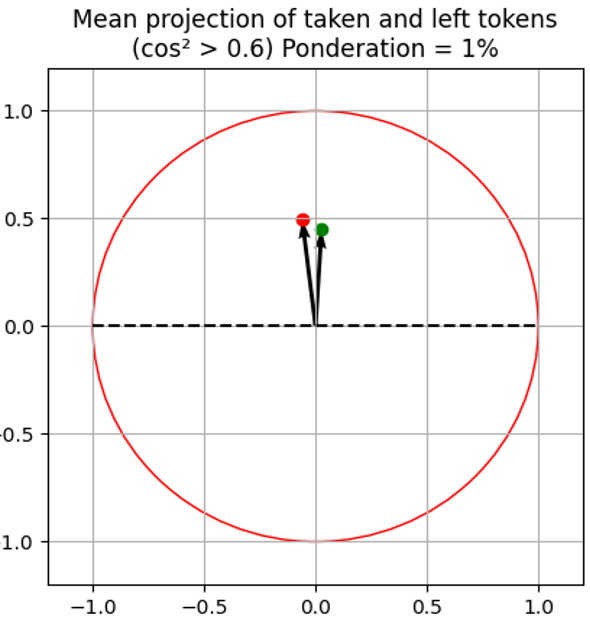}
    \captionsetup{justification=centering, font=small}
    \caption*{\textit{Graph n$^{\circ}$7: Reconstruction of PCA coordinates of the mean embedding vector associated with taken-tokens and the mean embedding vector associated with left-tokens (Weighting=1\% ; Layer 1 ; all cos$^{2} > .6$).}}
    \label{fig:graph7}
\end{figure}

More analytically, Table no. 10 highlights once again a minority of embedding variables associated with taken-tokens, consistently aligning with our earlier postulate of categorical selectivity, although this average minority (36\%) is stronger than in our previous illustrative cases. We observe the same type of results: scalar products, albeit weak for the aforementioned reasons, but still non-negligible, projecting the average embedding vectors onto the axis opposing taken and left-tokens.

\begin{table}[H]
    \centering
    \renewcommand{\arraystretch}{1}
    \small
    \rowcolors{1}{gray!20}{gray!40}
    \begin{tabular}{|c|c|c|}
        \hline
        & Taken-tokens & Left-tokens \\
        \hline
        Mean (\% projected variables) & 35.95 & 64.05 \\
        \hline
        Mean (mean(cos)) & .0639 & .166 \\
        \hline
        Mean (mean(scalar product)) & .0646 & .1177 \\
        \hline
    \end{tabular}
    \captionsetup{justification=centering, format=plain}
    \caption*{\textit{Table n$^{\circ}$10: Mean projection statistics of PCA coordinates of embedding variables on group membership axis (Weighting=1\% ; Layer 1 ; $N=950$ ; on the 79.62\% of variables with cos$>.6$; Where KMO $> 0.5$ ; Mean(KMO)=.501).}}
    \label{tab:table10}
\end{table}

The outlining genetically performed by the categories of target neurons involves abstracting categorical subdimensions from the categories of their precursor neurons, which are not categorically random but rather homogeneous or convergent. The projection of these singular subdimensions within the observational frame (the initial embedding space of GPT2-XL) allows us to highlight this partial categorical convergence, as evidenced by a segmentation and compartmentalization of these embedding dimensions: some dimensions, and only some, coalesce to constitute the categorical forms (the extracted subdimensions), which are distinct from a non-retained categorical background (the remaining embedding dimensions). However, the relatively weak correlations (between embedding dimensions and categorical subdimensions) observed is a particularly remarkable phenomenon: these subdimensions are not simply "extracted" in the sense of simple abstraction as defined by Piaget or philosophical empiricism. Their combinations are neither pre-given nor preexisting within the respective categorical dimensions of origin. On the contrary, and consistent with the ultimate stage of "reflection" within Piagetian synthetic reflective abstraction carried out by the aggregation function, outlining is the product of original and creative recombination of the initial embedding dimensions. Outlining is not merely the separation of a preexisting form from an existing background but rather the authentic and singular construction of a form, distinctly separated from a background that is just as fabricated.

\subsection{Outlining and Segmentation of Categorical Zones}
In the previous section, we attempted to highlight the fact that, within the framework of categorical outlining, the subdimensions extracted, or rather constructed, are each categorically convergent (and not random or semantically totally chaotic) at the internal level: each subdimension tends to be associated with specific initial embedding dimensions and not others. Let us now take a further step in understanding this categorical convergence. This, still within an approach of dimensional reduction of the vector space of the initial embeddings of GPT-2XL, but this time (i) by relaxing the constraint of a linear construction of the reduced categorical axes (thus taking into account the effect of the nonlinear activation function coupled with the aggregation function), (ii) by freeing ourselves from certain statistical conditions partially not respected in our previous PCA studies, and (iii) by coupling our approach with a token clustering approach (respectively taken and left). And still based on the (100) core-tokens of the precursor neurons' categories from layer 0, of which 15\% and 85\% become taken-tokens in the context of genetic interaction with a target neuron in layer 1.

Graph 8, created using the core-tokens of precursor neuron number 3537 from layer 0 and their taken (green points) or left (red points) status in the genetic interaction with target neuron number 10 in layer 1, is highly informative. Let us focus on the centroids of the taken-tokens (cf. dark green point labeled "Taken") and the left-tokens (cf. dark red point labeled "Left"); and take as a measure of distance whether these centroids are present in the same quadrants defined by the intersection of the two reduced dimensional axes obtained. We can observe that, although these centroids are not extremely far from each other, they are not present in the same quadrants, meaning they are not positioned in the same categorical zones of embeddings. Thus, the categorical subdimension extracted (or rather, created) here by the target neuron from the entire categorical dimension of the precursor neuron is indeed categorically defined: this subdimension (at its "Taken" centroid) belongs to a distinct categorical zone from that of the non-extracted categorical background (at its "Left" centroid). There is indeed, as with our previous results related to PCA, an abstraction of a relatively homogeneous and specific categorical subdimension in relation to an operationalized observation framework in terms of the initial embeddings of GPT-2XL (otherwise the centroids would be positioned in the same location and there would be no categorical segmentation). However, more qualitatively, made possible by the t-SNE approach which preserves local proximity structures between tokens, we can see that the involved taken-tokens ("reimburse," "bund," "enrolled," "subsidy," "insured," "deductible," "insurance," "coverage," "obamacare," etc.) tend to refer to a well-defined lexical field, here the lexical field of health insurance, a clearly distinct categorical form extracted from the categorical background formed by the left-tokens.

\begin{figure}[H]
    \centering
    \includegraphics[width=1\textwidth]{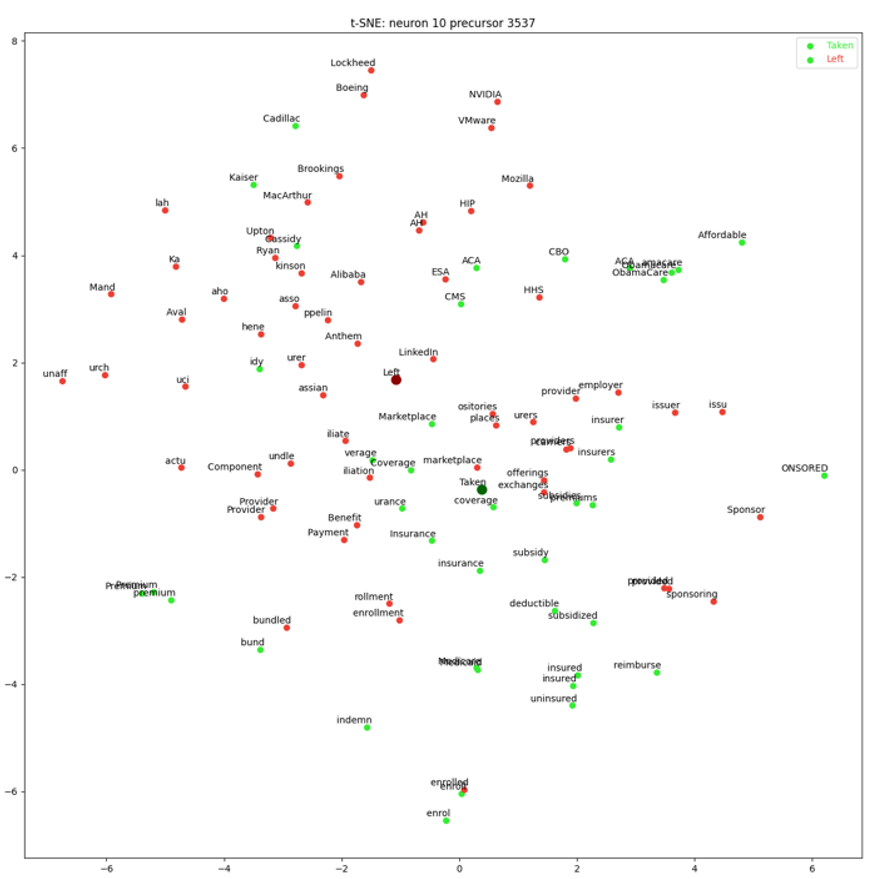}
    \captionsetup{justification=centering, font=small}
    \caption*{\textit{Graph n$^{\circ}$8: Distribution of taken and left-tokens in the embedding space reduced by t-SNE (Precursor neuron 3537 ; Destination neuron 10).}}
    \label{fig:graph8}
\end{figure}

The same applies to Graph 9, which relates to the precursor neuron pair 6010 / target neuron 93. The centroids are again positioned within contrasting categorical zones (quadrants). The taken-tokens ("wiki," "shares," "posted," "recap," "offline," "encyclopedia," "article," "emails," "subscribe," "folios," "tweet," "javascrip," "whats," etc.) that make up the categorical subdimension created here tend to converge around the lexical field related to digital communications and online platforms; this lexical field is relatively distinct from the semantics associated with the involved left-tokens.

\begin{figure}[H]
    \centering
    \includegraphics[width=1\textwidth]{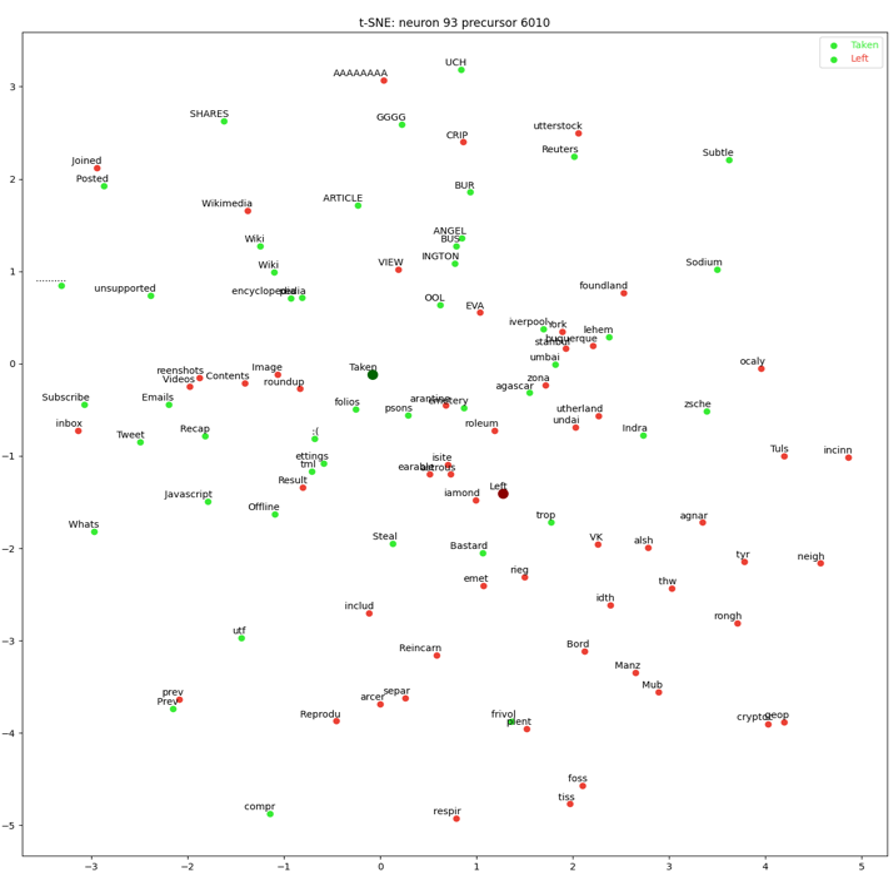}
    \captionsetup{justification=centering, font=small}
    \caption*{\textit{Graph n$^{\circ}$9: Distribution of taken and left tokens in the embedding space reduced by t-SNE (Precursor neuron 6010 ; Destination neuron 93).}}
    \label{fig:graph9}
\end{figure}

More systematically, we applied our t-SNE approach to the 1610 pairs (precursor neuron from layer 0, associated and strongly connected target neuron from layer 1) for which our t-SNE approach was possible and for which the percentage of taken-tokens extracted from the core-tokens ranged again between 15\% and 85\%. Graph 10 presents a synthetic reconstruction of the obtained data, showing the mean centroids of taken-tokens and left-tokens according to the different scenarios obtained from the crossing of their respective positions within the four quadrants (categorical embedding zones). The same trend, previously identified in the examples presented, is observed, and is highly significant ($\chi^2 = 217.75, p < .0001, \text{df} = 9$): the mean centroids of the extracted taken-tokens tend (with 87\% in this case) to be positioned within categorical zones (quadrants) different from those of the mean centroids of the associated left-tokens.

\begin{figure}[H]
    \centering
    \includegraphics[width=1\textwidth]{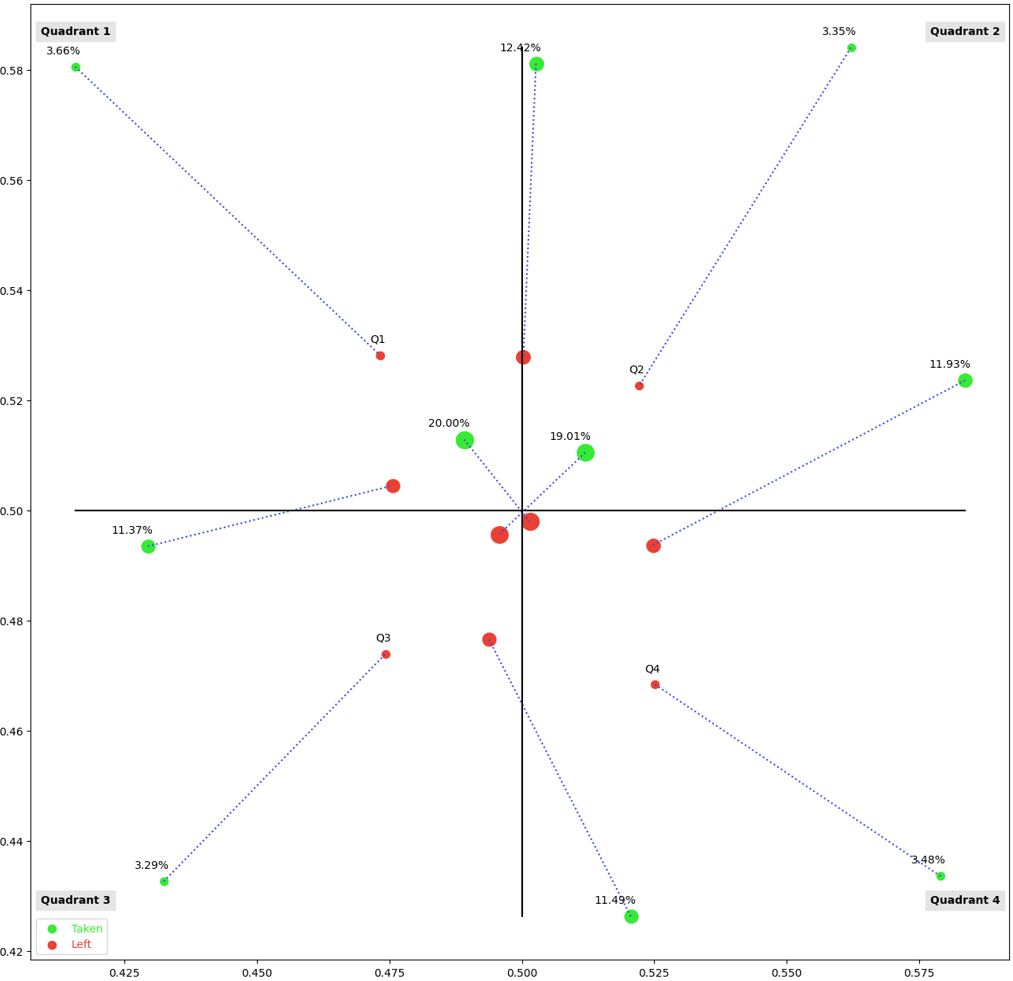}
    \captionsetup{justification=centering, font=small}
    \caption*{\textit{Graph n$^{\circ}$10: Graphical distribution, within quadrants, of mean centroids of paired taken and left-clusters in the embedding space reduced by mean t-SNE’s axes (Layers 0/1; $N=1610$).}}
    \label{fig:graph10}
\end{figure}

In line with the observations previously noted, Table n°11 demonstrates a systematic underrepresentation of associated taken/left barycenters positioned within the same quadrants, contrasted by a strong overrepresentation (39\% of the relevant cases) of associated barycenters dispersed between quadrants Q1 and Q4, or between quadrants Q2 and Q3. This latter point interestingly reveals that the extracted categorical subdimensions tend to be categorically distinct with respect to both t-SNE factorial axes jointly, thereby indicating an even greater degree of categorical distinctiveness.

\begin{figure}[H]
    \centering
    \includegraphics[width=0.9\textwidth]{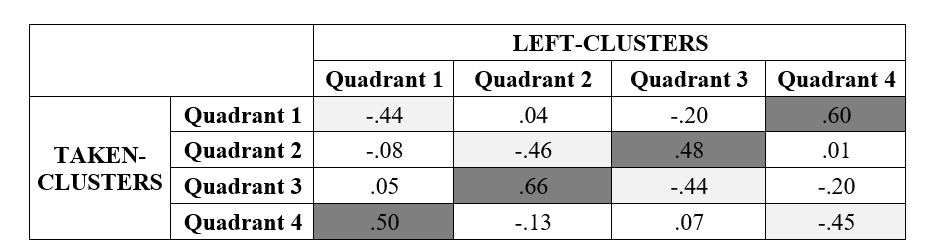}
    \captionsetup{justification=centering, font=small}
    \caption*{\textit{Table n$^{\circ}$11: Under- and over-representations of the distribution, within quadrants, of the mean centroids of paired taken and left-clusters in the embedding space reduced by mean t-SNE’s axes (Layers 0/1; $N=1610$).}}
    \label{fig:table11}
\end{figure}

The present t-SNE study reveals a convergence with the results of the previous PCA study: the categorical subdimensions extracted from the categories of precursor neurons (layer 0) by the aggregation functions of the associated and strongly connected target neurons (layer 1) are not categorically random but tend to be categorically homogeneous at the internal level, here from the perspective of the observation framework constituted by the vector space of GPT-2XL embeddings. More specifically, the categorical clipping performed from the category carried by a precursor neuron tends to create a categorical subdimension whose center of gravity is categorically distinct and (relatively) distant from the categorical center of gravity of the rest of this initial category. In other words, categorical clipping shapes, in categorically distinct zones, a categorical form (subdimension) actively distinguished from a categorical background. And a clustering approach of the (taken versus left) tokens involved reveals the qualitative semantic result of this process of disjunction, categorical differentiation performed by the phenomenon of categorical clipping (at least when the semantics involved corresponds to elements of human semantics, which can be assumed to be more strongly the case for the earlier layers).

\subsection{Categorical Phenomenologies of Clipping}

Let us conclude our current investigation of the characteristics of categorical clipping by taking a further step in its qualitative study. This will be done by presenting, purely as an example, a partial typology illustrating a possible way to categorize different categorical cases of clipping (see Table 12). These examples are all derived from a qualitative analysis of the characteristics of clusters of taken-tokens (the categorical form) extracted at layer 1 in relation to their corresponding clusters of left-tokens (the categorical background). All of these examples, in this case, follow a "human linguistic" approach.

A first categorical class of clipping can be described as "semantic." It can be subdivided into two subclasses:
\begin{itemize}
    \item First, hetero-lexical clipping, characterized by the segmentation of tokens belonging to the same lexical field, distinct from that of the categorical background:
    \begin{itemize}
        \item By splitting tokens with the same root. Example: the tokens "manager" and "managerial" are differentiated from a categorical background containing the tokens "Pharma," "Middles," "villa."
        \item By fragmentation of tokens from different roots. Example: the tokens "manager," "Wenger" (famous coach of the Arsenal football team), and "logistics" are dissociated from a categorical background including the tokens "Nurse," "Motorist," "frustrated."
    \end{itemize}
    \item Second, sub-lexical clipping, determined by the splitting of tokens paired with a sub-field lexical group within the lexical field of the involved category:
    \begin{itemize}
        \item By dispersion of tokens with the same root. Example: the tokens "order," "ordered," "ordering," "Ord" are separated from "want," "aspir," "needing," "desirable," "desire." Or: the tokens "Psych," "psychopath," "psychiatry" are distinguished from a category containing "McGill," "Graduation," "NYU," "lecturer," "academ," "Prof" (higher education lexical field) and "pediatric," "mindfulness," "PTSD," "hypnot," "clinicians" (medical lexical field).
        \item By disunion of tokens from distinct roots. Example: the tokens "listen," "Audio," "Speakers" (non-technical audio lexical field) are discriminated from the tokens "kHz," "dB," "Codec," "frequencies," "spectrum" (technical audio lexical field). Or: the tokens "audio," "listen," "ear," "sound," "headphone" (perceptual sound lexical field) are separated from the tokens "soundtrack," "Melody," "orchestra," "Musical," "Bach," "singers" (artistic sound production lexical field) and the tokens "dB," "MIDI," "kHz," "Frequency," "wav," "reverber" (technical sound lexical field), and even from the tokens "Yamaha," "drums," "violin," "bells," "guitar" (musical instruments lexical field).
    \end{itemize}
\end{itemize}

A second, more limited categorical class of clipping can be described as "graphemic." It can be characterized by the demarcation of tokens with a specific group of letters. Example: the tokens "ID," "id," "pid," "IDA" are isolated from "mom," "phone," "CAR," "but," "Ratio."

\begin{figure}[H]
    \centering
    \includegraphics[width=1\textwidth]{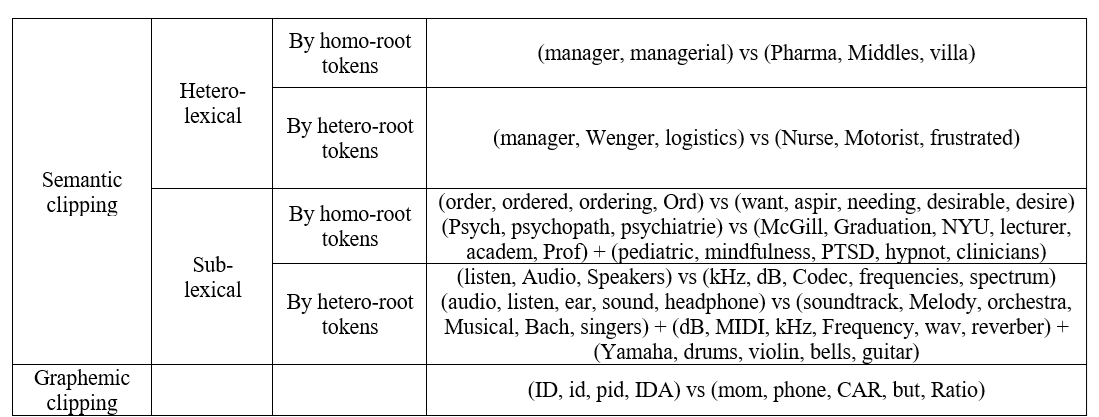}
    \captionsetup{justification=centering, font=small}
    \caption*{\textit{Table n$^{\circ}$12: Examples of types of linguistic categorial cut-outs : differences between left and taken-tokens from same neurons’ core-tokens (Layer 1).}}
    \label{fig:table12}
\end{figure}

These present examples, mentioned purely for illustrative purposes and without the intention of systematizing, help us qualitatively understand how categorical clipping can create extractions of categorical subdimensions, by grouping tokens that converge as part of a homogeneous and defined categorical segment. Even though these examples, related to early layers and thus still potentially close to human-like logic, are easily interpretable within the categories of human thought that we possess, we must not think that this is necessarily the case: the extracted subdimensions are often of the order of alien concepts, non-isomorphic to human primes of thought; there is indeed a rationality in their categorical convergence, but synthetic, that is, specific to statistical constructs that are not immediately or easily explainable within the categories of human semantics.

\section{Discussion}
\subsection{Synthesis of Our Exploratory Approach to the Synthetic Categorical Segmentation Process}

We have explored, in an exploratory manner, the process of categorical segmentation performed by the synthetic cognition of language models, consisting of cutting and creating new categorical dimensions within the world of tokens. Each formal MLP neuron can be associated with a specific categorical dimension, traceable through its own extension of afferent tokens that particularly activate for this category (its core-tokens).

From a causal structural perspective, this segmentation is, among other things, driven by the aggregation function inherent to each neuron \cite{Pichat2024d}, a function embodying three factors that govern the genesis and activation of categories carried by neurons in layer $n+1$ from the categories of precursor neurons in layer $n$: (i) the "x-effect" or synthetic categorical priming (the activation of precursor categories propagates and influences the creation of target categories), (ii) the "w-effect" or synthetic categorical attention (the connection weights between target and precursor neurons control the degree of relevance assigned to precursor categories in the construction of new target categories), and (iii) the "$\Sigma$-effect" or synthetic categorical phasing (subgroups of identical core-tokens from different precursors, simultaneously activated, reinforce each other categorically in the genesis of new target categories).

From a functional perspective, these three mathematical-cognitive factors of categorical segmentation govern, at the level of a target neuron (layer $n$), a mechanism for extracting a specific categorical subdimension from the category carried by each of its precursor neurons (layer $n-1$); the union of these subdimensions generates the new categorical dimension vectorized by this target neuron. This abstraction translates into a synthetic categorical clipping process, consisting of creating and distinguishing a form from a categorical background. The goal is to understand the properties of this categorical clipping, performed on the relative categorical variability of the tokens constituting the extension of each precursor neuron's category, in order to extract a subset of tokens that are categorically homogeneous and aligned with the (new) specific category created and carried by their corresponding target neuron.

Several functional characteristics of this categorical clipping have been proposed here in an exploratory manner:
\begin{itemize}
    \item \textbf{Categorical reduction}, related to the fact that the extracted categorical subdimension is associated with a set of (taken-)tokens that are categorically more homogeneous among themselves, compared to the set of (core-)tokens from the precursor category involved.
    \item \textbf{Categorical selectivity}, manifested by the extraction, from the category of a precursor neuron, of an extension quantitatively more restricted in terms of the number of (taken-)tokens.
    \item \textbf{Separation of initial embedding dimensions}, related to the fact that categorical clipping, when observed in the reference framework of the initial embedding vector space, tends to manifest as a dichotomous elective compartmentalization of these embeddings, with some being preferentially paired with the extracted categorical form (i.e., the subdimension), as opposed to others more associated with the remaining categorical background in the original category.
    \item \textbf{Segmentation of categorical zones}, manifested by a relative displacement of the categorical centers of gravity of the extracted form and the non-selected background, each of these centroids being positioned in contrasting categorical regions, attestable, when possible, by differentiated human semantic interpretations of the token clusters involved.
\end{itemize}

These different properties qualitatively give us an understanding of the various synthetic cognitive characteristics by which the categorical clipping process creates extractions of categorical subdimensions from precursor neurons, grouping tokens that converge into a form representing a homogeneous categorical segment that has been fabricated.

\subsection{What Epistemological Status to Attribute to the Subdimensions Extracted by Categorical Clipping?}

As we have already addressed in previous works \cite{Pichat2024a, Pichat2024e}, it is crucial not to fall into the epistemological trap of anthropomorphism, which involves attempting to analyze synthetic cognition solely through the filter of our own human cognitive and categorical concepts. It is important to understand that the categorical subdimensions extracted through categorical clipping are not necessarily aligned with conventional human categories of thought \cite{Fan2023, Bills2023, Olah2023, Olah2020, Bricken2023}.

Similarly, we must move beyond a form of epistemological naivety, both realistic and empiricist, that tends to "naturalize" the subdimensions instantiated by the categorical clipping process, believing that they represent subcategories that already exist in the material world and are corollaries of a form of intrinsic, pre-given, and ontological reality. As von Glaserfeld beautifully tells us \cite{Glaserfeld2002}: "In order to judge the goodness of a representation that is supposed to depict something else, one would have to compare it to what it is supposed to represent. In the case of 'knowledge' that would be impossible, because we have no access to the 'real' world except through experience and yet another act of knowing, and this, by definition, would simply yield another representation (...). It is logically impossible, however, to compare a representation with something it is supposed to depict, if that something is supposed to exist in a real world that lies beyond our experimental interface" (p.93). While the constructivist author indicates that the very idea of "'representation' (...) implies a reproduction, copy, or other structure that is in some way isomorphic with an original" (p.94), he invites us to prefer the formulation "re-presentation," which is more consistent with the idea that it denotes "a re-play of my own experiences, not a piece of some independent, objective world" (p.95).

\subsection{What Cognitive Status to Attribute to the Subdimensions Extracted by Categorical Clipping?}

The theory of conceptualization developed by Vergnaud \cite{Vergnaud2020a} in the field of human cognitive development seems particularly fruitful for providing us with a framework for thinking about the synthetic phenomena of categorical segmentation and categorical clipping construction.

Within Vergnaud's theory, conceptualization is a representational activity whose goal is the cognitive construction of operational characteristics; this is done in order to ground action on these characteristics and thus make it effective \cite{Vergnaud2020b}. In other words, the function of conceptualization is to establish homomorphisms between the realm of the objects in the world on which action is to be taken and the realm of operations and contents of thought. Thus, for Vergnaud \cite{Vergnaud2020c}, the process of conceptualization is fundamentally a pragmatic, economic cognitive activity: "one conceptualizes in order to act effectively." Moreover, to a large extent, for the author, conceptualization is not cognitively positioned within the realm of explicit, conscious, and verbalized theorization, but rather within the realm of action. This operational finality is at the root of the characteristic of conceptualization being "in act," meaning that it is encapsulated in action. Thus, the concepts and theorems mobilized by the individual are called "concepts-in-act" and "theorems-in-act" as they are only activated in action and serve no purpose other than to make the action effective.

Vergnaud \cite{Vergnaud2009} defines a concept-in-act as a category of thought deemed relevant by the individual in relation to a class of action situations. Concept-in-acts are categories of thought through which the subject creates and integrates information related to the type of situations they face. In other words, concept-in-acts are cognitive filters through which a given situation is "read" or constructed cognitively. From an epistemological perspective, there is potentially an infinite number of formal types of categories of thought; the most frequently encountered types are as follows: object, property, relation, transformation, condition, process. Again, concept-in-acts are pragmatic vectors of thought that organize information processing by segmenting objects in the world according to the contingent goals of the finalized activity \cite{Pichat2002}. Indeed, the functionality of concept-in-acts lies in the fact that they allow the subject to focus attention on a limited number of selected elements that are experienced as important for the success of the action. In this regard, they underlie a representation of the only situation variables considered central for the effectiveness of the action.

Based on these definitional elements, we can state that the category carried by a formal neuron exhibits, from an epistemological perspective, the defining traits of a concept-in-act. Indeed, a synthetic category is not a conscious, demonstrated, justified, or explained entity within synthetic cognition. Moreover, an artificial category fundamentally has a pragmatic purpose: its genesis and existence are finalized by the efficiency of information processing and the cognitive activity it enables: performing the tasks for which the neural network was designed and trained, encoding the context (positional and "semantic") of the tokens. We can therefore qualify the categories associated with formal neurons as synthetic concepts-in-act.

Vergnaud \cite{Vergnaud2016} defines a theorem-in-act as a proposition of thought deemed true by the subject in relation to a class of action situations. Theorems-in-act ground the effectiveness of action by resting on the implicit "practical theories" they constitute, which the subject develops from the interweaving of the properties of the objects they face. In other words, theorems-in-act allow the individual to build a practical representation of how the functional characteristics of action situations should be organized in order to act effectively upon them \cite{PichatMerri2007}. More specifically, these theorems-in-act underlie a representation of how to combine these critical situational variables. Formally, a proposition results from the composition of predicates and arguments; a theorem-in-act is thus an interweaving of concept-in-acts. In this regard, theorems-in-act structure activity in terms of instantiated properties, relations, conditions, and transformations. Pragmatically, theorems-in-act are practical propositions of thought from which action organizations (rules of action) derive. In this way, the rules of action are "operationalized corollaries" of theorems-in-act, with which they inherently maintain a certain morphic continuity.

As we have indicated previously, the activity of categorical segmentation performed by synthetic cognition is centrally regulated (among other things) by the defining aggregation function of each formal neuron. In its form, $\Sigma(w_{i,j} x_{i,j}) + b$, this aggregation function creates the category associated with a neuron in layer $n$ by specifically (weighted) combining the categories carried by these precursor neurons in layer $n-1$. In line with the previous definitions, an aggregation function can be assimilated to a synthetic theorem-in-act presiding over the singular coordination (specific to each neuron) of a series of preceding synthetic concept-in-acts. Such an artificial theorem-in-act effectively performs an activity of reflection, in the sense of Piaget’s reflective abstraction, as it carries out a genuine reconstruction and reorganization of precursor categories (in layer $n-1$), projected onto the higher plane of the new neural category generated in layer $n$; these initial categories being manipulated cognitively, in accordance with Piaget's definition, through the weighted sum vectorized by the aggregation function.

Let us take one final step further. The categorical subdimensions extracted from the categories of precursor neurons are themselves synthetic concepts-in-act; indeed, from an epistemological perspective, a subdimension is a category, or more precisely, a subcategory created within its original category. In this regard, a categorical subdimension can be referred to as a sub-concept-in-act.

But it is interesting to understand in what singular mathematical-cognitive context these synthetic concepts-in-act (i.e., categorical subdimensions) emerge: in the specific "paroxysmal" context where the theorem-in-act carried by the aggregation function is associated with a resulting activation value (i.e., the result of the calculation performed by the aggregation function) that is high. Indeed, by definition, the extracted categorical subdimensions are associated with taken-tokens, meaning tokens that are core-tokens of the target neuron, i.e., tokens with high activation in this target neuron. Now, by the mathematical construction of the aggregation function $\Sigma(w_{i,j} x_{i,j}) + b$, the output of this function is maximal when the three mathematical-cognitive factors associated with it—priming, attention, and categorical phasing—operate most strongly in tandem; that is, when simultaneously (i) the values of the connection weights $w_{i,j}$ are the highest, (ii) the activation values $x_{i,j}$ of the precursor categories are the highest, and (iii) the sum $\Sigma$ adds a maximum of cases where the product $w_{i,j} x_{i,j}$ itself is increased. It is only in this specific context that the categorical emergence of the sub-concepts-in-act (i.e., categorical subdimensions) extracted or rather fabricated by categorical clipping occurs in the proper sense. And it is in this paroxysmal context, which could actually be made visible numerically by relevant partial derivatives being zero, that the reflection stage of Piaget's reflective abstraction is fully realized: as the abstracted categories (precursor categories) are effectively projected and transformed onto a more abstract plane, resulting in subcategories (subdimensions) that are more ethereal; more abstract subcategories that will be combined to generate, properly speaking, the new category of the involved target neuron, subcategory by subcategory.

This is how we can understand how, within a form of categorical phase transition, the theorem-in-act defining the category of a target neuron clips sub-concepts-in-act (i.e., subdimensions) from the precursor categories and shapes this new categorical form, thus constructed and extracted from its original categorical background.

\section{Conclusion}
We have conducted an exploratory investigation in the field of neuropsychology of artificial intelligence, focusing on the modalities of categorical segmentation performed by a language model, specifically GPT-2XL. This process involves, through different neural layers, the creation of new categorical dimensions to analyze textual data and accomplish the tasks required by the model. In a multilayer perceptron (MLP) network, each neuron is associated with a specific category, determined by three factors derived from the neural aggregation function: priming, attention, and categorical phasing. At each new layer, these factors drive the emergence of new categories derived from the categories of the previous neurons. Through a process of categorical clipping, these new categories are formed by a selective abstraction of specific subdimensions of their antecedent categories, distinguishing a form from a categorical background. Several synthetic cognitive characteristics of this clipping have been identified here: categorical reduction, categorical selectivity, separation of initial embedding dimensions, and segmentation of categorical zones.

These properties of categorical clipping have been interpreted as manifestations of synthetic theorems-in-action, associated with neuronal aggregation functions, which, during a paroxysmal phase corresponding to the maxima of these functions, generate a reflective abstraction of singular synthetic sub-concepts-in-action. The recombination of these sub-concepts shapes the creation of new synthetic categories that are even more functional in relation to the activity objectives of the neural network involved.

In the context of a new study, currently underway, we continue our exploration of synthetic categorical segmentation by attempting to better understand how categorical restructuring occurs from one neural categorical layer $n$ to its subsequent layer $n+1$. This is done by investigating the synthetic cognitive phenomenology through which sub-concepts-in-act (i.e., categorical subdimensions), clipped from different precursor neurons, are either semantically and activationally convergent with one another, thus generating, in their associated target neurons, new original and singular categorical structures of synthetic cognition.

\section*{Acknowledgments}
The authors would like to thank Madeleine Pichat for her careful review of this article, the ER IPC research team (Free Faculties of Philosophy and Psychology of Paris), within which the "Neocognition" sub-team carried out this study, Andrew Ponomarev (Petersburg Federal Research Center of the Russian Academy of Sciences) for his stimulating insights, Judicael Poumay (Neocognition) for his valuable insights, and Jeanne-Théoline Reybier (Chrysippe) for her personal activities that made the conditions for conducting this study possible.

\end{document}